\documentclass[letterpaper]{article}
\usepackage{aaai2026}
\usepackage{times}
\usepackage{helvet}
\usepackage{courier}
\usepackage[hyphens]{url}
\usepackage{graphicx}
\usepackage{natbib}
\usepackage{caption}
\usepackage{algorithm}
\usepackage{algorithmic}
\usepackage{array}
\usepackage{amsmath}
\usepackage{amsfonts}

\usepackage{pgfplotstable}
\pgfplotsset{compat=1.18}

\usepackage{booktabs}
\usepackage{siunitx}
\usepackage{longtable}
\usepackage{booktabs}

\usepackage{newfloat}
\usepackage{listings}
\usepackage{arydshln}

\DeclareCaptionStyle{ruled}{labelfont=normalfont,labelsep=colon,strut=off}
\floatstyle{ruled}
\newfloat{listing}{tb}{lst}{}
\floatname{listing}{Listing}
\usepackage{epsfig}
\usepackage{soul}
\usepackage{soul,xcolor}
\usepackage{tabularx}
\newcolumntype{R}{>{\raggedleft\arraybackslash}X}

\usepackage[capitalize]{cleveref}
\usepackage{array}
\newcommand{\ours}{\textsc{Dabgo}}
\newcommand{\trackstar}{\textsc{TrackStar}}
\newcommand{\bm}{\textsc{Bm25}}
\newcommand{\Gecko}{\textsc{Gecko}}
\nocopyright

\definecolor{hlA}{RGB}{90,130,200}
\definecolor{hlB}{RGB}{95,134,203}
\definecolor{hlC}{RGB}{100,138,206}
\definecolor{hlD}{RGB}{105,142,209}
\definecolor{hlE}{RGB}{110,146,212}
\definecolor{hlF}{RGB}{115,150,215}
\definecolor{hlG}{RGB}{120,154,218}
\definecolor{hlH}{RGB}{125,158,221}
\definecolor{hlI}{RGB}{130,162,224}
\definecolor{hlJ}{RGB}{135,166,227}
\definecolor{hlK}{RGB}{140,170,230}
\definecolor{hlL}{RGB}{145,174,233}
\definecolor{hlM}{RGB}{150,178,236}
\definecolor{hlN}{RGB}{155,182,239}
\definecolor{hlO}{RGB}{160,186,242}
\definecolor{hlP}{RGB}{165,190,244}
\definecolor{hlQ}{RGB}{170,194,246}
\definecolor{hlR}{RGB}{175,198,248}
\definecolor{hlS}{RGB}{180,202,250}
\definecolor{hlT}{RGB}{185,206,251}
\definecolor{hlU}{RGB}{190,210,252}
\definecolor{hlV}{RGB}{195,214,253}
\definecolor{hlW}{RGB}{200,218,254}
\definecolor{hlX}{RGB}{205,222,254}
\definecolor{hlY}{RGB}{210,226,254}
\definecolor{hlZ}{RGB}{215,230,255}
\definecolor{hlAA}{RGB}{220,235,255}
\definecolor{hlAB}{RGB}{230,240,255}
\definecolor{hlAC}{RGB}{240,246,255}
\definecolor{hlAD}{RGB}{248,253,255}

\newcommand{\hlA}[1]{\sethlcolor{hlA}\hl{#1}}
\newcommand{\hlB}[1]{\sethlcolor{hlB}\hl{#1}}
\newcommand{\hlC}[1]{\sethlcolor{hlC}\hl{#1}}
\newcommand{\hlD}[1]{\sethlcolor{hlD}\hl{#1}}
\newcommand{\hlE}[1]{\sethlcolor{hlE}\hl{#1}}
\newcommand{\hlF}[1]{\sethlcolor{hlF}\hl{#1}}
\newcommand{\hlG}[1]{\sethlcolor{hlG}\hl{#1}}
\newcommand{\hlH}[1]{\sethlcolor{hlH}\hl{#1}}
\newcommand{\hlI}[1]{\sethlcolor{hlI}\hl{#1}}
\newcommand{\hlJ}[1]{\sethlcolor{hlJ}\hl{#1}}
\newcommand{\hlK}[1]{\sethlcolor{hlK}\hl{#1}}
\newcommand{\hlL}[1]{\sethlcolor{hlL}\hl{#1}}
\newcommand{\hlM}[1]{\sethlcolor{hlM}\hl{#1}}
\newcommand{\hlN}[1]{\sethlcolor{hlN}\hl{#1}}
\newcommand{\hlO}[1]{\sethlcolor{hlO}\hl{#1}}
\newcommand{\hlP}[1]{\sethlcolor{hlP}\hl{#1}}
\newcommand{\hlQ}[1]{\sethlcolor{hlQ}\hl{#1}}
\newcommand{\hlR}[1]{\sethlcolor{hlR}\hl{#1}}
\newcommand{\hlS}[1]{\sethlcolor{hlS}\hl{#1}}
\newcommand{\hlT}[1]{\sethlcolor{hlT}\hl{#1}}
\newcommand{\hlU}[1]{\sethlcolor{hlU}\hl{#1}}
\newcommand{\hlV}[1]{\sethlcolor{hlV}\hl{#1}}
\newcommand{\hlW}[1]{\sethlcolor{hlW}\hl{#1}}
\newcommand{\hlX}[1]{\sethlcolor{hlX}\hl{#1}}
\newcommand{\hlY}[1]{\sethlcolor{hlY}\hl{#1}}

\title{Data Attribution in Large Language Models\\via Bidirectional Gradient Optimization}
\author{
    Frédéric Berdoz,
    Luca A. Lanzendörfer,
    Kaan Bayraktar,
    Roger Wattenhofer
}
\affiliations{
    ETH Zurich\\
    \{fberdoz, lanzendoerfer, kbayraktar, wattenhofer\}@ethz.ch

}

\begin{document}

\maketitle

\insert\footins{\noindent\footnotesize Presented at the AI Governance (AIGOV) Workshop at AAAI 2026.}

\begin{abstract}
Large Language Models (LLMs) are increasingly deployed across diverse applications, raising critical questions for governance, accountability, and data provenance. Understanding which training data most influenced a model's output remains a fundamental open problem. We address this challenge through training data attribution (TDA) for auto-regressive LLMs by expanding upon the inverse formulation: How would training data be affected if the model had seen the generated output during training? Our method perturbs the base model using bidirectional gradient optimization (gradient ascent and descent) on a generated text sample and measures the resulting change in loss across training samples. Our framework supports attribution at arbitrary data granularity, enabling both factual and stylistic attribution. We evaluate our method against baselines on pretrained models with known datasets, and show that it outperforms previous work on influence metrics, thereby enhancing model interpretability, an essential requirement for accountable AI systems.
\end{abstract}

\begin{links}
\link{Code}{https://github.com/ETH-DISCO/DABGO}
\end{links}

\section{Introduction}
Large language models (LLMs) have emerged as transformative tools capable of performing a wide range of tasks, from drafting legal documents \cite{siino2025exploring}, providing medical diagnoses \cite{zhou2025large}, to assisting scientific research \cite{luo2025llm4sr}.
Although LLMs are increasingly equipped with external tools to search the Internet and retrieve information,
their value remains primarily rooted in their knowledge and creativity acquired during training on vast text corpora.
Consequently, current research primarily focuses on developing more effective models given a dataset, whereas the inverse problem of identifying which part of that dataset was most influential given an LLM output remains underexplored, despite its potential to enable traceable and auditable AI systems. Besides improving interpretability, such capabilities would close the feedback loop between training data and model behavior, enabling a wide range of applications, such as model debugging and unlearning \cite{tanno2022repairing} and data valuation \cite{sim2022data}. 
We introduce \textbf{D}ata \textbf{A}ttribution through \textbf{B}idirectional \textbf{G}radient \textbf{O}ptimization (\ours{}). This framework estimates data influence by comparing the training loss between two models optimized with respect to the generated text, one model optimized via gradient ascent and the other model optimized via gradient descent. This allows influence to be attributed at any level of data granularity.
To validate \ours{}, we conduct experiments on open-ended text generation in both factual and stylistic settings, demonstrating that our approach outperforms prior attribution methods. In summary, our contributions are as follows:

\begin{itemize}
\item We introduce \ours, a simple attribution method for open-ended text generation using auto-regressive LLMs.
\item We quantitatively demonstrate that our approach significantly outperforms recent attribution baselines.
\item We qualitatively demonstrate that \ours{} captures both factual content and stylistic characteristics in attributed texts, leveraging the interpretability of our method.
\end{itemize}

\begin{figure*}[t]
  \centering
  \includegraphics[width=\textwidth]{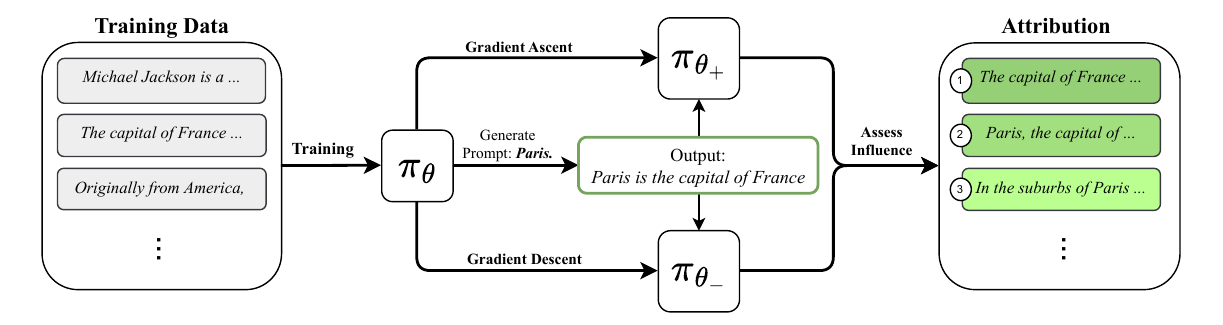}  
  \caption{\textbf{Overview of \ours{}.} We propose a bidirectional attribution technique for training data attribution in LLMs. Starting with a model $\pi_\theta$ trained from scratch, we generate an output sentence from a minimal prompt (e.g., “Paris.”). We then apply bidirectional gradient optimization (descent and ascent) on the generated output to obtain two optimized models $\pi_{\theta_-}$ and $\pi_{\theta_+}$, respectively. To assess the influence of each training sample, we compute its loss under both optimized models and rank samples by the absolute change in loss. This yields a ranking of the most influential training samples responsible for the generated text.}
  \label{fig:attribution}
\end{figure*}

\section{Related Work}

\subsection{Training Data Attribution}
Training data attribution (TDA) quantifies how much a given training example influenced a model’s prediction on a test input. The gold-standard definition is the counterfactual effect of removing a sample from the training set, but exact computation requires retraining for each sample \citep{cook1977detection} or averaging over all subsets \citep{ghorbani2019data} and is therefore infeasible. A wide range of different approaches have been proposed to estimate the influence of samples without retraining, such as datamodels \citep{ilyas2022datamodels}, simulators of alternative training runs \citep{guu2023simfluence} and checkpoint based influence estimation \citep{pruthi2020estimating}.
Most modern TDA methods rely on the concept of \emph{influence functions} from classical robust statistics \cite{hampel1974influence}. Influence functions estimate the effect of an infinitesimal change in the weight of any training sample and can be computed in closed form via a Hessian-adjusted dot product between model gradients of the training and test samples \cite{koh2017understanding}. Although promising, \citet{schioppa2023theoretical} mention some practical and fundamental limitations of traditional influence function estimation.

\subsection{Forward Influence Function Estimation}
Due to the high memory and compute cost associated with inverting the Hessian matrix of a model, most modern TDA methods introduce efficient approximation techniques. \cite{schioppa2022scaling, arnoldi1951principle, agarwal2017second, park2023trak, grosse2023studying, kwon2024datainf, chang2025scalable}. 
Among these, we primarily compare to \citet{chang2025scalable}, who demonstrated promising performance in pretraining data attribution. Their method, \trackstar, utilizes randomly projected gradients to approximate the Hessian.

\subsection{Backward Influence Function Estimation}
A recent and promising line of work exploits \emph{the mirrored influence hypothesis} formulated by \citet{ko2024mirrored}, which states that the influence of a training sample over a model prediction can be accurately estimated by the influence of that same model prediction over the original training example. Reverse influence is typically estimated by unlearning the model on the generated output and measuring the change in training loss between the original and unlearned models. To focus on the parameters most critical to the generated output, the unlearning step is usually performed using a Fisher-regularized gradient ascent step, a method known as machine unlearning \cite{bourtoule2021machine}. This approach has the advantage of not using training sample gradients, which greatly improves scalability \cite{ko2024mirrored}, and has been applied in various modalities, including text \cite{isonuma2024unlearning}, vision \cite{wang2024data}, and audio \cite{choi2025large}.
However, prior natural language processing (NLP) studies either restrict themselves to fact tracing using curated facts with single-word predictions \cite{ko2024mirrored} or operate only at dataset-level granularity \cite{isonuma2024unlearning}. In contrast, we adapt and extend this line of work to fully open-ended text generation, and we introduce a key refinement: we perform both Fisher-regularized gradient ascent and descent, which we find improves attribution.

\section{Methodology}
We propose a framework for estimating backward influence functions that enables tracing generated model outputs back to their most influential training samples. Let $\mathcal{D} = \{x^i\}_{i=1}^N$ be the training dataset, where each $x^i=x^i_0x^i_1...x^i_{L-1} \in \mathcal{X}^L$ represents a contiguous segment of text in the training corpus, $L$ the context window used during training, and $\mathcal{X}$ the token vocabulary.
Our method assumes an auto-regressive language model parameterized by $\theta$, denoted $\pi_\theta$. Given a prompt $x_0$ of length $l_0$ (we simplify and slightly abuse notation by representing the entire prompt as a single token $x_0$) and a partial completion $x_1...x_{t-1}$, the model predicts a probability distribution $\pi_\theta(x_t | x_{0:t-1})$ over $\mathcal{X}$, which can then be used at inference to sample a continuation $x_t$. With this notation, the likelihood of any sequence $x=x_1...x_l$ given prompt $x_0$ can be expressed as
\begin{equation}
\label{eq:autoregressive}
\pi_\theta(x) = \prod_{t=1}^l \pi_\theta(x_t|x_{0:t}),
\end{equation}
and the negative log-likelihood loss incurred by $\pi_\theta$ on $x$ as
\begin{equation}
\label{eq:loss}
\ell(x, \theta) = -\frac{1}{l}\sum_{t=1}^{l}\log \pi_\theta(x_t|x_{0:t-1}).
\end{equation}
Finally, we denote by $\pi_{\theta^*}$ the pretrained model from which we aim to perform TDA, where
\begin{equation}
\label{eq:pretraining}
\theta^* \approx \arg\min_{\theta} \frac{1}{N}\sum_{i=1}^N \ell(x^i, \theta).
\end{equation}

\subsection{Backward Influence Estimation}
To assess which training samples most strongly influenced a particular model completion $\hat{x}_{1:l}$ on a user-defined prompt $\hat{x}_0$, we take inspiration from \citet{ko2024mirrored} and reverse the traditional perspective of influence functions. Rather than analyzing the effect of training samples on model outputs (forward influence), we study how optimizing the model on the test sample $\hat{x}$ affects its behavior on the training data $\mathcal{D}$ (backward influence).
 We construct two perturbed variants by optimizing on the test sample $\hat{x}=\hat{x}_0\hat{x}_1...\hat{x}_l$ using two symmetric procedures: gradient ascent and gradient descent, yielding two new sets of model weights, $\theta^{\hat{x}}_+$ and $\theta^{\hat{x}}_-$, respectively. Specifically, similar to \citet{wang2024data}, we initialize $\smash{\theta_\pm^{(0)} = \theta^*}$ and choose $\smash{\theta^{\hat{x}}_\pm = \theta_\pm^{(M)}}$ for some $M$, where 
\begin{equation}
    \label{eq:finetuning}
    \theta_\pm^{(m+1)} \leftarrow \theta_\pm^{(m)} \pm \frac{\alpha}{N}F_{\theta^*}^{-1}\nabla \ell(x, \theta),
\end{equation}
and with $\alpha$ a learning rate and $F_{\theta^*}$ the Fisher Information Matrix (FIM), given by
\begin{equation}
\label{eq:exactFIM}
F_{\theta^*} := \left.\mathbb{E}_{x \sim \mathcal{D}} \left[ \left.\nabla_\theta \log \pi_\theta(x)\right|_{\theta^*} \nabla_\theta \log \pi_\theta(x)\right|_{\theta^*}^\top \right] .
\end{equation}
Projecting the gradient using the inverse FIM is known as \emph{Elastic Weight Consolidation} (EWC), a method designed to mitigate catastrophic forgetting \cite{kirkpatrick2017overcoming}. Since inverting the exact FIM is computationally intractable, we follow \citet{choi2025large} and use the diagonal approximation
\begin{equation}
\label{eq:approxFIM}
 \left(\hat{F}_{\theta^*}\right)_{jj}= \frac{1}{NL}\sum_{x^i \in \mathcal{D}} \sum_{t=1}^L   \left(\left.\frac{\partial  \log \pi_\theta(x^i_t|x^i_{0:t-1})}{\partial \theta }\right|_{\theta = \theta^*}\right)^2 
\end{equation}
With the perturbed models in hand, we compute the loss of each training sample $x^i \in \mathcal{D}$ under both $\theta^{\hat{x}}_-$ and $\theta^{\hat{x}}_+$. Finally, we define the bidirectional influence score (BIS) of $x^i$ towards the test sample $\hat{x}$ as the absolute change in loss:
\begin{equation}
\label{eq:influence_score}
\mathcal{I}(x^i;\hat{x}) = |\ell(x^i,\theta^{\hat{x}}_- ) - \ell(x^i, \theta^{\hat{x}}_+)|.
\end{equation}
Training examples that exhibit large changes in loss under these two models can be interpreted as those potentially most affected by the test sample and, by the \emph{mirrored influence hypothesis} \cite{ko2024mirrored}, most influential in the generation of $\hat{x}$. Our method, \emph{Data Attribution via Bidirectional Gradient Optimization} (\ours{}) ranks training samples in descending order of their BIS, enabling targeted inspection and interpretation of the data underlying specific model outputs. A diagram illustrating our method is provided in \cref{fig:attribution}.

\section{Experiments}
\begin{table*}[!t]
\small
\centering
\setlength{\tabcolsep}{0mm}
\begin{tabularx}{\linewidth}{p{1.6cm} RRRRRRR p{0.4cm} RRRRRRR}
\toprule
 & \multicolumn{7}{c}{Wikipedia} & &\multicolumn{7}{c}{Gutenberg} \\
 \midrule
 \textbf{Method}& 1 & 3 & 5 & 7 & 10 & 15 & 20 & &1 & 3 & 5 & 7 & 10 & 15 & 20 \\
\midrule
Random & 1.6 & 4.3 & 4.7 & 6.3 & 8.8 & 8.7 & 8.9 & & 42.9 & 46.5 & 45.6 & 45.3 & 50.0 & 53.5 & 42.8 \\
\bm{} & \textbf{159.1} & \underline{90.8} & \underline{67.3} & \underline{59.4} & \underline{57.0} & \underline{68.7} & \underline{68.3} & &75.7 & 90.3 & 95.0 & 110.2 & 103.0 & 103.5 & 104.4 \\
\trackstar{} & 3.9 & 6.3 & 7.4 & 8.5 & 8.9 & 9.7 & 10.3 & & 61.7 & 57.2 & 50.8 & 51.4 & 56.2 & 54.8 & 60.6 \\
Gecko & 43.5 & 55.6 & 55.2 & 42.1 & 29.9 & 18.2 & 14.5 & & \textbf{127.5} & \textbf{143.9} & \underline{177.7} & \textbf{291.4} & \underline{215.8} & \underline{312.4} & \underline{258.5} \\

\ours{} & \underline{133.8} & \textbf{101.2} & \textbf{75.9} & \textbf{73.9} & \textbf{77.7} & \textbf{89.3} & \textbf{76.7} & &\underline{98.3} & \underline{112.2} & \textbf{258.6} & \underline{235.6} & \textbf{355.1} & \textbf{377.0} & \textbf{470.7} \\

\bottomrule
\end{tabularx}
\caption{\textbf{Quantitative analysis of \ours{} against baselines.} We report the tail-patch absolute scores for varying numbers of top-$k$ proponents, as defined in \cref{eq:tpa}. Bold indicates the best-performing method, and underlining marks the second best. Results are shown for both factual (Wikipedia) and stylistic (Gutenberg) attribution. In the factual setting, \ours{} consistently outperforms all baselines across values of $k$, except at $k=1$. Here \bm{} also identifies a relevant training sample. This is likely due to strong lexical overlap between generated and training sequences in the factual setting, which \bm{} is well-suited to capture. In the stylistic setting, our method performs competitively with \Gecko{} at small $k$ and outperforms as $k$ increases, reflecting that stylistic influence is typically distributed across multiple training samples rather than a single passage.}
\label{tab:tail-patch-results}
\end{table*}

We evaluate \ours{} on two language models based on the GPT-2 architecture \cite{radford2019language}, each trained from scratch on a distinct corpus to support both factual and stylistic attribution. For each generated query $\hat{x}$ we perform $M=10$ gradient descent and ascent steps following \cref{eq:finetuning}, with learning rate $\alpha=1 \cdot 10^{-4}$.

\subsection{Pretraining}
We evaluate \ours{} in two complementary settings: factual attribution, using a curated collection of Wikipedia abstracts, and stylistic attribution, using literary texts from the Project Gutenberg archive \cite{projectgutenberg2025}. The Wikipedia corpus (230M tokens) is derived from the WIT dataset \cite{srinivasan2021wit} and provides a mostly uniform writing style, forcing attribution to rely on factual content rather than stylistic signal. In contrast, the Gutenberg corpus (2M tokens) introduces meaningful stylistic variation across authors and time periods, enabling attribution of stylistic influence.
In both cases, we train a GPT-2–style model from scratch with a context window of 256 tokens and create overlapping training blocks using a sliding window of stride 128.
Preprocessing includes tokenization, removal of boilerplate, and text normalization.

\subsection{Evaluation}
We evaluate our attribution method by quantifying the extent to which the top-$k$ attributed training samples influence the likelihood assigned by the model to a given generated test sample.

\paragraph{Tail-Patch Absolute Score.}
Unlike fact-tracing settings, open-ended generation does not provide a tractable ground truth for attribution: generated text may combine factual, stylistic, or compositional features that do not correspond to any single training example. This makes retrieval- or entailment-based metrics unsuitable, as they assume access to a gold reference passage. Moreover, surface-level overlap is not the objective of training data attribution, which seeks causal influence rather than lexical similarity. 
Instead, we adopt the \emph{tail-patch absolute} (TPA) evaluation protocol, which quantifies the additive influence of a set of training samples on a given test sample \cite{chang2025scalable}. TPA offers a computationally efficient proxy for exact influence estimation, which involves retraining from scratch without the attributed samples and is therefore intractable in most practical settings.
The evaluation proceeds as follows: for each generated query sentence $\hat{x}$, we identify the top-$k$ most influential training samples using the attribution method under consideration. These samples are then used to perform a single gradient update step on the base model $\pi_{\theta^*}$, resulting in a perturbed model with parameters $\theta_k$. We compute the likelihood of $\hat{x}$ under both $\pi_{\theta^*}$ and $\pi_{\theta_k}$, and define the TPA metric $\tau_k$ as the absolute change in likelihood, that is
\begin{equation}
\label{eq:tpa}
\tau_k = \frac{|\pi_{\theta^*}(\hat{x}) -\pi_{\theta_k}(\hat{x})|}{\pi_{\theta^*}(\hat{x})}.
\end{equation}
This formulation treats attribution as a sensitivity analysis. It is based on the hypothesis that influential training samples should perturb the model's likelihood of $\hat{x}$ more strongly than non-influential samples, either positively or negatively. Using the absolute difference captures the overall coupling between training and test samples. This is particularly important in open-ended generation settings, where influence is not necessarily unidirectional.
We observe that computing $\tau_k$ on $k$ random training samples yields the lowest values across all tested attribution methods (see \cref{tab:tail-patch-results}), suggesting that the metric is robust to noise and reflects meaningful influence rather than random variation.

\begin{figure}[t] 
  \centering
  \includegraphics[width=\columnwidth]{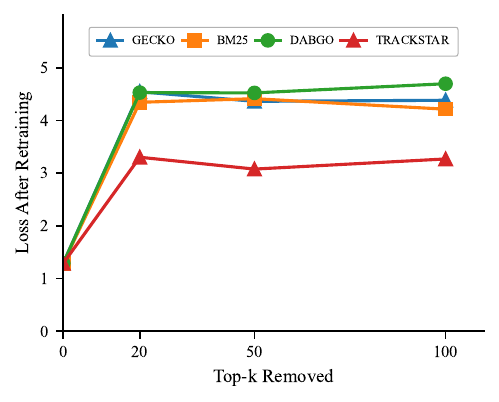}
  \caption{\textbf{Loss after retraining without top-$k$.} We plot the average final loss across four generated sentences $\hat{x}$ from five distinct Gutenberg queries, after removing their top-$k$ proponents from the training set and retraining the model. The resulting increase in loss provides strong evidence that our method accurately identifies influential training samples. Moreover, the alignment between this empirical effect and our tail patch absolute scores supports the validity of our main evaluation metric.}
  \label{fig:retrained-losses}
\end{figure}

\paragraph{Retraining from Scratch.}
To validate our main evaluation metric, the tail-patch score, we perform a counterfactual evaluation via full retraining. For each attribution method (\bm{}, \trackstar{}, \Gecko{} and \ours{}), we identify the top-$k$ influential training samples $\mathcal{D}_k^{\hat{x}} \subset \mathcal{D}$ for a given query $\hat{x}$. We then remove these samples to construct a reduced training set $\mathcal{D}_{-k}^{\hat{x}} = \mathcal{D} \setminus \mathcal{D}_k^{\hat{x}}$, and retrain a new model $\pi_{\theta_{-k}}$ on this subset. We keep all training hyperparameters identical to the base model $\pi_\theta$.
To quantify the effect of removal, we compare the loss of the test sample $\hat{x}$ under the retrained model, $\ell(\hat{x}, \theta_{-k})$, to its original loss under the base model, $\ell(\hat{x}, \theta)$. A higher value of $\ell(\hat{x}, \theta_{-k})$ indicates greater influence of removed training samples, validating the effectiveness of the attribution method.
We perform this experiment on four test samples from the Gutenberg models with $k = 20, 50, 100$, corresponding to $0.1\%$, $0.3\%$, and $0.6\%$ of the full training set. As shown in \cref{fig:retrained-losses}, models retrained without samples attributed by \ours{} consistently yield higher losses compared to those trained without samples from \bm{}, \trackstar{} or \Gecko{}. While computationally expensive, this procedure corresponds to the ground-truth definition of influence and provides validation that the tail-patch score is a meaningful and efficient evaluation method.

\paragraph{Comparison Baselines.}
To evaluate the quality of the identified influential samples, we compare our method against several baselines. These include \trackstar{}~\citep{chang2025scalable}, which estimates influence via per-sample loss gradients, Best Matching 25 (\bm{}) \citep{robertson1994some}, a classical term-based retrieval method that ranks documents (sequences in our case) by estimating their relevance to a query, and \Gecko{} \citep{lee2024gecko}, a text-embedding based retrieval method. Similar to the authors in the open-ended text generation experiment \citep{chang2025scalable}, we do not use the query-specific Hessian matrix approximation for \trackstar{}. 
We also include random training samples as a noise baseline to contextualize the inherent variability in $\tau_k$. Evaluation is based on the average absolute change in probability across multiple query samples.

\subsection{Factual Attribution}
\paragraph{Generation.} 
We sample subject entities from the Wikipedia dataset~\cite{srinivasan2021wit} and use them to construct generation prompts. For each subject $s$, we prompt the model with $\hat{x}_0=s$. This simple prompting strategy minimizes prompt-induced bias, while enabling targeted generation that reduces the number of relevant samples to attribute, thereby facilitating both qualitative and quantitative evaluation of \ours{}. We generate continuations using Top-$50$ sampling with temperature $1$ and a repetition penalty of $1.5$. This setup produces standalone sentences or factual statements about the subject. Examples of our prompts and continuations can be found in \cref{app:prompting_strategy}.

\paragraph{Attribution.} Unlike previous work, which often attributes human-written answers to handcrafted prompts (e.g., attributing why the model assigns probability to \textit{``France''} when prompted with \textit{``Paris is in''}), we perform attribution on full model-generated sentences, making it both more realistic and more challenging. We compute $\tau_k$ over 25 query samples across varying values of $k$ and present our results in \cref{tab:tail-patch-results}. We observe that our proposed method \ours{} yields consistently stronger attribution performance than evaluated baselines. 
We further conduct ablations isolating the effect of gradient ascent (unlearning-style updates used in prior work \cite{choi2025large, wang2024data, isonuma2024unlearning}) and descent (finetuning-style updates). While ascent alone already yields strong performance, combining both ascent and descent, leads to the highest attribution accuracy (cf.\ \cref{tab:tail-patch-ablation}). We attribute this to the generative nature of LLMs: model outputs differ in how “likely” they are by the underlying training distribution, and ascent and descent capture complementary signals. A qualitative examples is provided \cref{app:qualitative_examples_appendix}.

\begin{table}[ht]
\setlength{\tabcolsep}{1mm}
\centering
\small
\begin{tabularx}{\columnwidth}{p{2.1cm}
>{\centering\arraybackslash}p{0.75cm}
>{\centering\arraybackslash}p{0.75cm}
>{\centering\arraybackslash}X
>{\centering\arraybackslash}X
>{\centering\arraybackslash}X
>{\centering\arraybackslash}X
>{\centering\arraybackslash}X}
\toprule
\textbf{Method} & \multicolumn{7}{c}{\textbf{Tail-patch absolute [\%] for different $k$}} \\
\midrule
  & 1 & 3 & 5 & 7 & 10 & 15 & 20  \\
\midrule
Descent & 16.6 & 22.5 & 26.4 & 28.8 & 32.4 & 37.4 & 40.3 \\
Ascent & 131.6 & 83.4 & 58.2 & 67.7 & 59.3 & 58.9 & 58.5  \\
\ours{} & \textbf{133.8} & \textbf{101.2} & \textbf{75.9} & \textbf{73.9} & \textbf{77.7} & \textbf{89.3} & \textbf{76.7} \\
\bottomrule
\end{tabularx}
\caption{\textbf{Ablation study on uni-directional attribution methods} using the Wikipedia dataset. We compare the top-$k$ proponents identified by three different influence estimation methods: Ascent-only: $\ell(\cdot, \theta) - \ell(\cdot,\theta_+)$, Descent-only: $\ell(\cdot, \theta_-) - \ell(\cdot, \theta)$, and \ours{}. For each method, we report the tail-patch absolute scores $\tau_k$, averaged over 25 sequences generated from distinct prompts.}
\label{tab:tail-patch-ablation}
\end{table}

\subsection{Stylistic Attribution}

\paragraph{Generation.} In this experiment, the model is prompted with a short sentence sampled from a book, and a continuation is generated (in this case, we do not use a repetition penalty to avoid stylistic bias). More specifically, for each author, we select a validation chunk not seen during training and extract the first complete sentence from that chunk to use as a prompt. We filter for prompts of moderate length to ensure the resulting generation remains within the model context window and has reasonable length. Although no ground-truth attribution targets are available, we make use of metadata associated with the prompt (e.g., author and book title) to assume that influential training samples should exhibit stylistic similarity to the origin of the source. To construct the evaluation queries, we sample one segment of text from each author represented in the subset.  
\paragraph{Attribution.} Quantitative results for the model trained on the subset are reported in the right half of \cref{tab:tail-patch-results}. \ours{} consistently outperforms \bm{} and \trackstar{} in this setting. The stylistic attribution setting shows the limitations of word-based retrieval methods such as \bm{} and the improvement in attribution with \Gecko{}. This is also observable in the qualitative results (cf.~\cref{tab:qualitative}), where we observe that \ours{} attributes to a sample originating from the same author as the segment of text used in the prompt, while also more accurately covering the topic and narrative which was generated.

\subsection{Interpretability}
Beyond ranking training samples, our method supports attribution at arbitrary levels of granularity. In particular, we refine our influence computation to the token level by measuring the absolute difference in per-token log-likelihoods under the updated models. Specifically, for the token at position $t$ in sample $x^i$, we compute
\begin{equation}
\label{eq:token_score}
\sigma(x^i_t;\hat{x})=|\log \pi_{\theta_+^{\hat{x}}}(x^i_{t}|x^i_{0:t-1}) - \log \pi_{\theta_-^{\hat{x}}}(x^i_{t}|x^i_{0:t-1})|.
\end{equation} 
This enables finer-grained interpretability by localizing which parts of a training sample contribute most to its attribution score. In Table~\ref{tab:qualitative}, we show examples from both factual and stylistic settings, highlighting the most influential tokens within each attributed training sample. This approach is especially beneficial for stylistic attribution, where subtle differences in sentence structure, syntax, and phrasing play a key role. For instance, the stylistic example shown in \cref{tab:qualitative} emphasizes first-person narratives, while the factual example highlights complete descriptive segments like “centered on the city of Rome.”

\subsection{Limitations and Future Work}
\ours{} remains computationally expensive: it requires two full passes over the training corpus, which is impractical for industrial-scale LLMs trained on hundreds of billions of tokens \cite{llama3_card}. Accordingly, our experiments are limited to controlled settings.
Although \ours{} supports arbitrary granularity, it attributes influence at the individual-sample level and does not account for interactions between samples. Subgroups of examples may exert synergistic or adversarial influence that is not recoverable from per-sample estimates. While \citet{isonuma2024unlearning} consider attribution at the dataset level, subgroup-level attribution remains an open problem and is even more computationally demanding.
Finally, while the highlighted tokens in attributed passages offer qualitative interpretability, we do not establish their causal role. Token-level attribution remains unverified beyond observed loss changes, and isolating true causal components is an important direction for future work.

\section{Conclusion}

We present \ours{}, a novel framework for training data attribution in autoregressive language models, based on backward influence estimation. By comparing training loss under models optimized via gradient ascent and descent, our method identifies influential training samples without expensive per-sample gradient computation. \ours{} supports attribution at arbitrary granularity, applies to both factual and stylistic outputs, and enables fine-grained interpretability analysis, providing transparency critical for accountable AI systems. Empirical results demonstrate that \ours{} qualitatively and quantitatively outperforms baselines on multiple attribution tasks. Additional experiments with models retrained from scratch further validate its effectiveness, showing strong agreement with ground-truth influence metrics. Unlike prior approaches that rely on handcrafted prompts and static completions, \ours{} can attribute in fully open-ended generation settings, offering a practical tool for interpretable and responsible deployment of LLMs.
{\small
\bibliography{references}
}

\appendix

\onecolumn 
\section{Prompting Strategy}
\label{app:prompting_strategy}
In \cref{tab:prompting_strategy_examples} we show some examples of our prompting strategy for our models and the types of outputs we received. 
\begin{table*}[ht]
\centering
\small
\begin{tabularx}{\linewidth}{p{2cm}X}
\toprule
\textbf{Prompt} & \textbf{Generated Text} \\
\midrule
Mount Everest. & Mount Everest is the highest point in the Himalayas at 8{,}948 m above sea level. \\
\hline
World War I. & It was a global military conflict that embroiled most of the world's great powers, assembled in two opposing alliances: the Entente and the Central Powers. The immediate cause of the war was the June 28, 1914 assassination of Archduke Franz Ferdinand, heir to the Austro-Hungarian throne, by Gavrilo Princip, a Bosnian Serb citizen of Austria–Hungary and member of the Black Hand. \\
\hline
Art Deco. & Art Deco is a style of visual arts, architecture and design that first appeared in France just before World War I. \\
\bottomrule
\end{tabularx}
\caption{\textbf{Prompting Strategy.} We show three examples of model outputs for subject-only prompts of the form \textit{``[subject].''}. The model tends to generate standalone factual statements guided by the topic, without relying on a relational or narrative context. While minor factual errors may occur (e.g., Everest's height is off by 100m), this does not undermine the attribution setting.}
\label{tab:prompting_strategy_examples}
\end{table*}

\section{Additional Qualitative Examples}
\label{app:qualitative_examples_appendix}

 \cref{tab:qualitative} shows factual attribution example from Wikipedia and a stylistic attribution example from Project Gutenberg.

\begin{table*}[!th]
\centering
{\fontsize{5}{6}\selectfont
\begin{tabularx}{\linewidth}{p{0.6cm}X}
\toprule
\textbf{Method} & \textbf{Top Proponent} 
\\
\midrule
&\textbf{Model}: GPT-2 trained on a subset of Wikipedia abstracts
\\ 
&\textbf{Prompt}: \textit{Ancient Rome.}
\\
& \textbf{Generated}: \textit{It was a civilization of ancient Rome centered on the city of Rome and its surrounding region.}
\\
\midrule
\bm{} & \textbf{From Wikipedia article: \emph{Culture of ancient Rome}} \newline  
The culture of ancient Rome existed throughout the almost 1200-year history of the civilization of Ancient Rome. The term refers to the culture of the Roman Republic, later the Roman Empire, which at its peak covered an area from Lowland Scotland and Morocco to the Euphrates. Life in ancient Rome revolved around the city of Rome, its famed seven hills, and its monumental architecture such as the Colosseum, Trajan’s Forum, and the Pantheon. The city also had several theaters, gymnasia, and many taverns, baths, and brothels. Throughout the territory under ancient Rome’s control, residential architecture ranged from very modest houses to country villas, and in the capital city of Rome, there were imperial residences on the elegant Palatine Hill, from which the word palace is derived. The vast majority of the population lived in the city center, packed into insulae. The city of Rome was the largest megalopolis of that time, with a population that may well have exceeded one million people, with a high-end estimate of 3.6 million and a low-end estimate of 450,000.
\\
\midrule
\trackstar{}  & \textbf{From Wikipedia article: \emph{List of rulers in the British Isles}} \newline 
This is a list of rulers in the British Isles. In 1603, King James VI of Scotland also became James I of England, joining the crowns of England and Scotland in personal union. By royal proclamation, James styled himself “King of Great Britain”, but no such kingdom was actually created until 1707, when England and Scotland united to form the new Kingdom of Great Britain, with a single British parliament sitting at Westminster, during the reign of Queen Anne. 
\\
\midrule
\Gecko{} & \textbf{From Wikipedia article: \emph{Campaign history of the Roman military}} \newline
From its origin as a city-state on the peninsula of Italy in the 8th century BC, to its rise as an empire covering much of Southern Europe, Western Europe, Near East and North Africa to its fall in the 5th century AD, the political history of Ancient Rome was closely entwined with its military history. The core of the campaign history of the Roman military is an aggregate of different accounts of the Roman military's land battles, from its initial defense against and subsequent conquest of the city's hilltop neighbors on the Italian peninsula, to the ultimate struggle of the Western Roman Empire for its existence against invading Huns, Vandals and Germanic tribes. These accounts were written by various authors throughout and after the history of the Empire. Following the First Punic War, naval battles were less significant than land battles to the military history of Rome due to its encompassment of lands of the periphery and its unchallenged dominance of the Mediterranean Sea.
The Roman army battled first against its tribal neighbours and Etruscan towns within Italy, and later came to dominate the Mediterranean and at its height the provinces of Britannia and Asia Minor.\\
\midrule
\textbf{\ours{} \newline (ours)} &
\textbf{From Wikipedia article: \emph{Outline of Ancient Rome}} \newline 
The following outline is provided as an overview of and topical guide to ancient\hlK{ Rome}:
Ancient Rome – former\hlH{ civilization}\hlI{ that} thrived on the\hlM{ Italian} Peninsula as early as the 8th century BC. Located\hlJ{ along} the\hlF{ Mediterranean} Sea and\hlE{ centered}\hlD{ on} the\hlC{ city}\hlA{ of}\hlB{ Rome},\hlG{ it} expanded to\hlL{ become} one of the largest empires in the\hlO{ ancient} world.
\\
\midrule
\midrule
&\textbf{Model}: GPT-2 trained on Project Gutenberg archive
\\
 & \textbf{Prompt (from Edward P. Roe's \textit{His Sombre Rivals})}: \textit{On reaching Centerville, we found everything in the direst confusion.}
\\
&  \textbf{Generated}: \textit{As we succeeded in taking it a letter from us to which we saw the enemy. I lowered the flank of our way, for the time that the enemy had given me a roar of fare, though not only was there. Your men had left me, but I could not help feeling that, at length, I saw a soldier among them, and that he would be sure to be shut out the march. As it was, I was quite sure, that he could not bring himself to any possible excuse for a surgeon. I could not have missed my life since}
 \\
 \midrule
 \bm{} & \textbf{From Alice I. Jones's \textit{Unveiling a Parallel: A Romance}} \newline
the performance was quite as distasteful to him as to me, the only difference being that he was familiar with the idea and I was not. As I passed out, I observed that many of the women were vaporizing and many of the men smoking. I suppose it was, in part, the intolerable abomination of these \\
\midrule
\trackstar{} & \textbf{From J.\ Berg Esenwein's \textit{The Art of Public Speaking}} \newline
both; it may be physical or spiritual, or both the mental image is at once the most lawless and the most law-abiding child that has ever been born of the mind. First of all, as its name suggests, the process of imagination—for we are thinking of it now as a process rather than as a faculty—is memory at work. Therefore we must consider it primarily as 1.~Reproductive Imagination. We see or hear or feel or taste or smell something and the sensation passes away. Yet we are conscious of a greater or lesser ability to reproduce such feelings at will. Two considerations, in general, will govern the vividness \\
\midrule

\Gecko{} & \textbf{From Edward P. Roe's \textit{His Sombre Rivals}} \newline
 could judge, was the position of affairs between twelve and one, although I can give you only my impressions. It appeared to me that our men were fighting well, gradually and steadily advancing, and closing in upon the enemy. Still, I cannot help feeling that if we had followed up our success by the determined charge of one brigade that would hold together, the hill might have been swept, and victory made certain. "I had taken my position near Rickett's and Griffin's batteries on the right of our line, and decided to follow them up, not only because they were doing splendid work, but also for the reason that they would\\
 \midrule

\textbf{\ours{} \newline(ours)} & \textbf{From Edward P. Roe's \textit{His Sombre Rivals}} \newline
Be\hlH{a}\hlP{ure}\hlN{gard},\hlR{ but}\hlV{ also} Johnson from the\hlK{ Shen}andoah. "My hope was exceedingly intensified by the appearance of a long line of troops emerging from the woods on our flank\hlJ{ and} rear, for\hlQ{ I} never dreamed that they could be other than our own re-\hlU{en}forcements. Suddenly I\hlG{ caught} sight of a\hlM{ flag} which\hlO{ I}\hlT{ had}\hlE{ learned} to know too well. The line halted a moment, muskets were\hlY{ leve}lled, and I found myself in a perfect storm of\hlW{ bullets}. I assure you I made a rapid change of base, for when our line turned\hlF{ I}\hlL{ should}\hlD{ be} between two\hlX{ fires}.\hlB{ As} it\hlA{ was}\hlI{,}\hlC{ I} was\hlS{ cut} twice \\
\bottomrule
\end{tabularx}}
\caption{\textbf{Qualitative comparison for a factual (top) and a stylistic (bottom) attribution example.} We show the top proponent retrieved by \ours{} and other baseline methods. Highlighted tokens correspond to the largest loss differences in \ours{}, as defined in \cref{eq:token_score}. In the factual case, the model is prompted with \textit{``Ancient Rome.''} and generates a sentence containing \textit{``centered on the city of Rome,''} which \ours{} successfully attributes to a training sample containing this exact phrase (also highlighted as the most influential segment), from the Wikipedia article \textit{Outline of Ancient Rome}. While \bm{} also retrieves a relevant training sample, \trackstar{} does not manage to surface semantically meaningful content.
In the stylistic example, \ours{} and \Gecko{} are the only methods that retrieves a thematically consistent, first-person battlefield narrative. In \ours{} (several \textit{``I''} are highlighted along with words such as \textit{``flag''}, \textit{``fires''}, \textit{``bullets''}). In addition, \ours{} returns as the main proponent a passage from the same author and book as the prompt segment used to generate the completion.}
\label{tab:qualitative}
\end{table*}

\end{document}